\documentclass[runningheads]{llncs}
\usepackage{graphicx}

\usepackage{tikz}
\usepackage{comment}
\usepackage{amsmath,amssymb} 
\usepackage{color}

\usepackage{subfigure}
\usepackage{booktabs}
\usepackage{multirow}
\usepackage{makecell}
\usepackage{booktabs}
\usepackage{colortbl}
\usepackage{xcolor}
\usepackage{algorithmicx,algorithm}
\usepackage[colorlinks,linkcolor=red,anchorcolor=red,citecolor=green]{hyperref}

\usepackage[accsupp]{axessibility}  

\begin{document}

\title{Masked Generative Distillation} 

\makeatletter
\newcommand{\printfnsymbol}[1]{%
  \textsuperscript{\@fnsymbol{#1}}%
}

\titlerunning{Masked Generative Distillation}
%
\author{Zhendong Yang\textsuperscript{$\star$}\textsuperscript{$\dagger$}\inst{1,2} \and
Zhe Li\textsuperscript{$\dagger$}\inst{2} \and
Mingqi Shao\inst{1} \and Dachuan Shi\inst{1}\\ Zehuan Yuan\inst{2}\and Chun Yuan\textsuperscript{$\ddagger$}\inst{1}}

\authorrunning{Z. Yang et al.}
%
\institute{$^{1}$Tsinghua Shenzhen International Graduate School \quad
$^{2}$ByteDance Inc\\
\email{\{yangzd21,smq21,sdc21\}@mails.tsinghua.edu.cn}\quad
\email{axel.li@outlook.com}\quad
\email{yuanzehuan@bytedance.com}\quad
\email{yuanc@sz.tsinghua.edu.cn}}
\maketitle
\renewcommand{\thefootnote}{\fnsymbol{footnote}} 
\footnotetext[1]{This work was performed while Zhendong worked as an intern at ByteDance.}
\footnotetext[4]{Equal Contribution} 
\footnotetext[5]{Corresponding author} 

\begin{abstract}
Knowledge distillation has been applied to various tasks successfully. The current distillation algorithm usually improves students' performance by imitating the output of the teacher. This paper shows that teachers can also improve students' representation power by guiding students' feature recovery. From this point of view, we propose Masked Generative Distillation (MGD), which is simple: we mask random pixels of the student's feature and force it to generate the teacher's full feature through a simple block. MGD is a truly general feature-based distillation method, which can be utilized on various tasks, including image classification, object detection, semantic segmentation and instance segmentation. We experiment on different models with extensive datasets and the results show that all the students achieve excellent improvements. Notably, we boost ResNet-18 from 69.90\% to 71.69\% ImageNet top-1 accuracy, RetinaNet with ResNet-50 backbone from 37.4 to 41.0 Boundingbox mAP, SOLO based on ResNet-50 from 33.1 to 36.2 Mask mAP and DeepLabV3 based on ResNet-18 from 73.20 to 76.02 mIoU. Our codes are available at \url{https://github.com/yzd-v/MGD}.

\keywords{Knowledge distillation, image classification, object detection, semantic segmentation, instance segmentation}
\end{abstract}

\section{Introduction}
Deep Convolutional Neural Networks (CNNs) have been widely applied to various computer vision tasks. Generally, a larger model has a better performance but a lower inference speed, making it hard to be deployed with a limited source. To get over this, knowledge distillation has been proposed.\cite{hinton2015distilling}. It can be divided into two types according to the location of distillation. The first is specially designed for the different tasks, such as logit-based distillation\cite{hinton2015distilling,zhou2020rethinking} for classification and head-based distillation\cite{dai2021general,zhixing2021distilling} for detection. The second is feature-based distillation\cite{romero2014fitnets,heo2019comprehensive,chen2021distilling}. Since only the head or projector after the feature is different among various networks, theoretically, the feature-based distillation method can be used in various tasks. However, distillation methods designed for a specific task are often unavailable for other tasks. For example, OFD\cite{heo2019comprehensive} and KR\cite{chen2021distilling} bring limited improvement for detectors. FKD\cite{zhang2020improve} and FGD\cite{yang2021focal}, specifically designed for detectors, cannot be utilized in other tasks due to the lack of neck.

The previous feature-based distillation methods usually make students imitate the teacher's output as closely as possible because the teacher's feature has a stronger representation power. However, we believe that it is unnecessary to mimic the teacher directly to improve the representation power of students' features. The features used for distillation are generally high-order semantic information through deep networks. The feature pixels already contain the information of adjacent pixels to a certain extent. Therefore, if we can use partial pixels to restore the teacher's full feature through a simple block, the representational power of these used pixels can also be improved. From this point of view, we present \textbf{M}asked \textbf{G}enerative \textbf{D}istillation (MGD), which is a simple and efficient feature-based distillation method. As shown in Figure~\ref{fig:architecture}, we first mask random pixels of the student's feature and then generate the teacher's full feature with the masked feature through a simple block. Since random pixels are used in each iteration, all pixels will be used throughout the training process, which means the feature will be more robust and its representation power of it will be improved. In our method, the teacher only serves as a guide for students to restore features and does not require the student to imitate it directly.

\begin{figure}[!ht]
\centering
\includegraphics[{width=0.86\linewidth}]{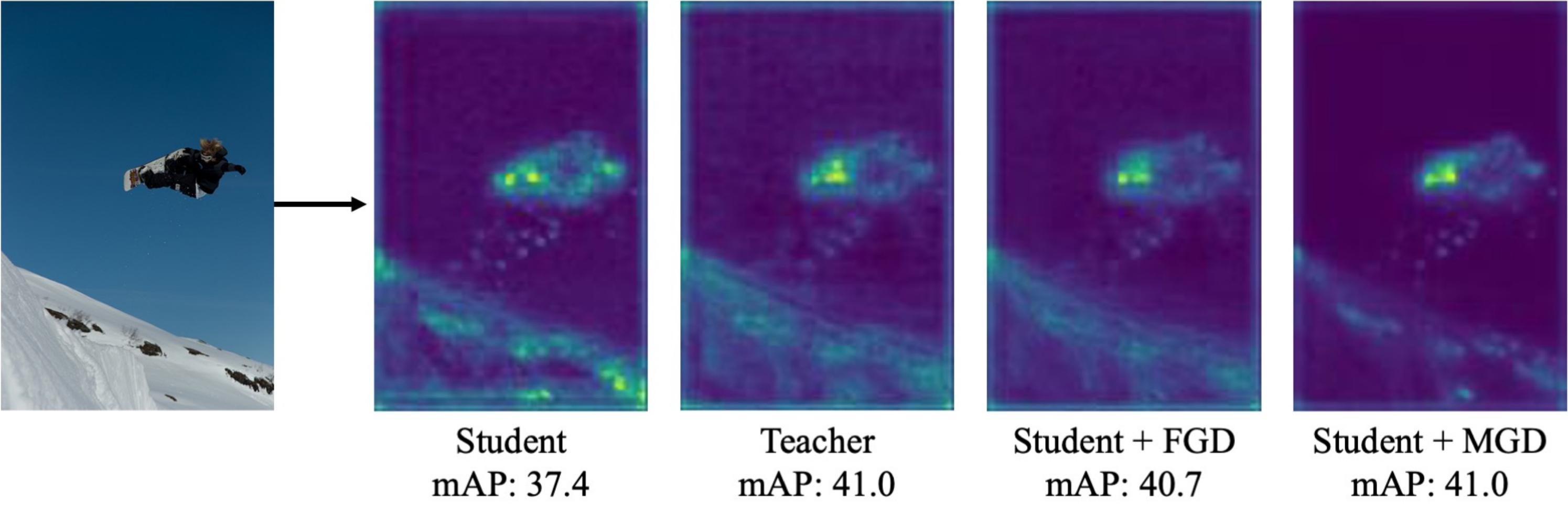}
\caption{Visualization of the feature from the first layer of FPN outputs. {\bf Teacher}: RetinaNet-ResNeXt101. {\bf Student}: RetinaNet-ResNet50. FGD\cite{yang2021focal} is a distillation method for detectors which forces the student to mimic the teacher's feature.}
\label{fig:visualization}
\end{figure}

In order to confirm our hypothesis that without directly imitating the teacher, masked feature generation can improve students' feature representation power, we do the visualization of the feature attention from student's and teacher's neck. As Figure~\ref{fig:visualization} shows, the features of student and teacher are quite different. Compared with the teacher, the background of the student's feature has higher responses. The teacher's mAP is also significantly higher than the student's, 41.0 vs. 37.4. After distillation with a state-of-the-art distillation method FGD\cite{yang2021focal}, which forces the student to mimic the teacher's feature with attention, the student's feature becomes more similar to the teacher's, and the mAP is greatly improved to 40.7. While after training with MGD, there is still a significant difference between the feature of the student and teacher, but the response to the background of the student is greatly reduced. We are also surprised that the student's performance exceeds FGD and even reaches the same mAP as the teacher. This also shows that training with MGD can improve the representation power of students' features. Besides, we also do abundant experiments on image classification and dense prediction tasks. The results show that MGD can bring considerable improvement to various tasks, including image classification, object detection, semantic segmentation and instance segmentation. MGD can also be combined with other logit-based or head-based distillation methods for even greater performance gains. To sum up, the contributions of this paper are:

\begin{enumerate}
  \item
  We introduce a new way for feature-based knowledge distillation, which makes the student generate the teacher's feature with its masked feature instead of mimicking it directly.

  \item

  We propose a novel feature-based distillation method, Masked Generative Distillation, which is simple and easy to use with only two hyper-parameters.

  \item
  We verify the effectiveness of our method on various models via extensive experiments on different datasets. For both image classification and dense prediction tasks, the students achieve significant improvements with MGD.
  
\end{enumerate}

\section{Related work}
\subsection{Knowledge Distillation for Classification}
Knowledge distillation was first proposed by Hinton et al.\cite{hinton2015distilling}, where the student is supervised by the labels and the soft labels from the teacher's last linear layer. However, more distillation methods are based on the feature map besides logit. FitNet\cite{romero2014fitnets} distills the semantic information from the intermediate layer. AT\cite{zagoruyko2016paying} summaries the values across the channel dimension and transfers the attention knowledge to the student. OFD\cite{heo2019comprehensive} proposes margin ReLU and designs a new function to measure the distance for distillation. CRD\cite{tian2019contrastive} utilizes contrastive learning to transfer the knowledge to students. More recently, KR\cite{chen2021distilling} builds a review mechanism and utilizes multi-level information for distillation. SRRL\cite{yang2020knowledge} decouples representation learning and classification, utilizing the teacher's classifier to train the student's penultimate layer feature. WSLD\cite{zhou2020rethinking} proposes the weighted soft labels for distillation from a perspective of bias-variance trade-off.

\subsection{Knowledge Distillation for Dense Prediction}
There is a big difference between classification and dense prediction. Many distillation works for classification have failed on dense prediction. Theoretically, the feature-based distillation method should be helpful for both classification and dense prediction tasks, which is also the goal of our method.

\subsubsection{Knowledge Distillation for object detection.} Chen et al.\cite{chen2017learning} first calculate the distillation loss on the detector's neck and head. The key to distillation for object detection is where to distill due to the extreme imbalance between foreground and background. To avoid introducing noise from the background, FGFI\cite{wang2019distilling} utilizes the fine-grained mask to distill the regions near objects. However, Defeat\cite{guo2021distilling} points out that information from foreground and background are both important. GID\cite{dai2021general} chooses the areas where the student and teacher perform differently for distillation. FKD\cite{zhang2020improve} uses the sum of the teacher's and student's attention to make the student focus on changeable areas. FGD\cite{yang2021focal} proposes focal distillation which forces the student to learn the teacher's crucial parts and global distillation which compensates for missing global information.

\subsubsection{Knowledge Distillation for semantic segmentation.} Liu et al.\cite{liu2019structured} propose pair-wise and holistic distillation, enforcing pair-wise and high-order consistency
between the outputs of the student and teacher. He et al.\cite{he2019knowledge} reinterpret the output from the teacher network to a re-represented latent domain and capture long-term dependencies from the teacher network. CWD\cite{shu2021channel} minimizes the Kullback–Leibler (KL) divergence between the probability map which is calculated by normalizing the activation map of each channel.

\section{Method}

The architectures of models vary greatly for different tasks. Moreover, most distillation methods are designed for specific tasks. However, the feature-based distillation can be applied to both classification and dense prediction. The basic method for distillation on features can be formulated as:
\begin{equation}
    L_{fea}=\sum_{k=1}^{C}\sum_{i=1}^{H}\sum_{j=1}^{W}\big( F_{k,i,j}^{T}-f_{align}(F_{k,i,j}^{S})\big)^{2}
  \label{general_feature_loss}
\end{equation}
where $F^{T}$ and $F^{S}$ denote the teacher's and student's feature, respectively, and $f_{align}$ is the adaptation layer to align student's feature $F^{S}$ with teacher's feature $F^{T}$. $C, H, W$ denotes the shape of the feature map.

\begin{figure}[!ht]
\centering
\includegraphics[{width=0.82\linewidth}]{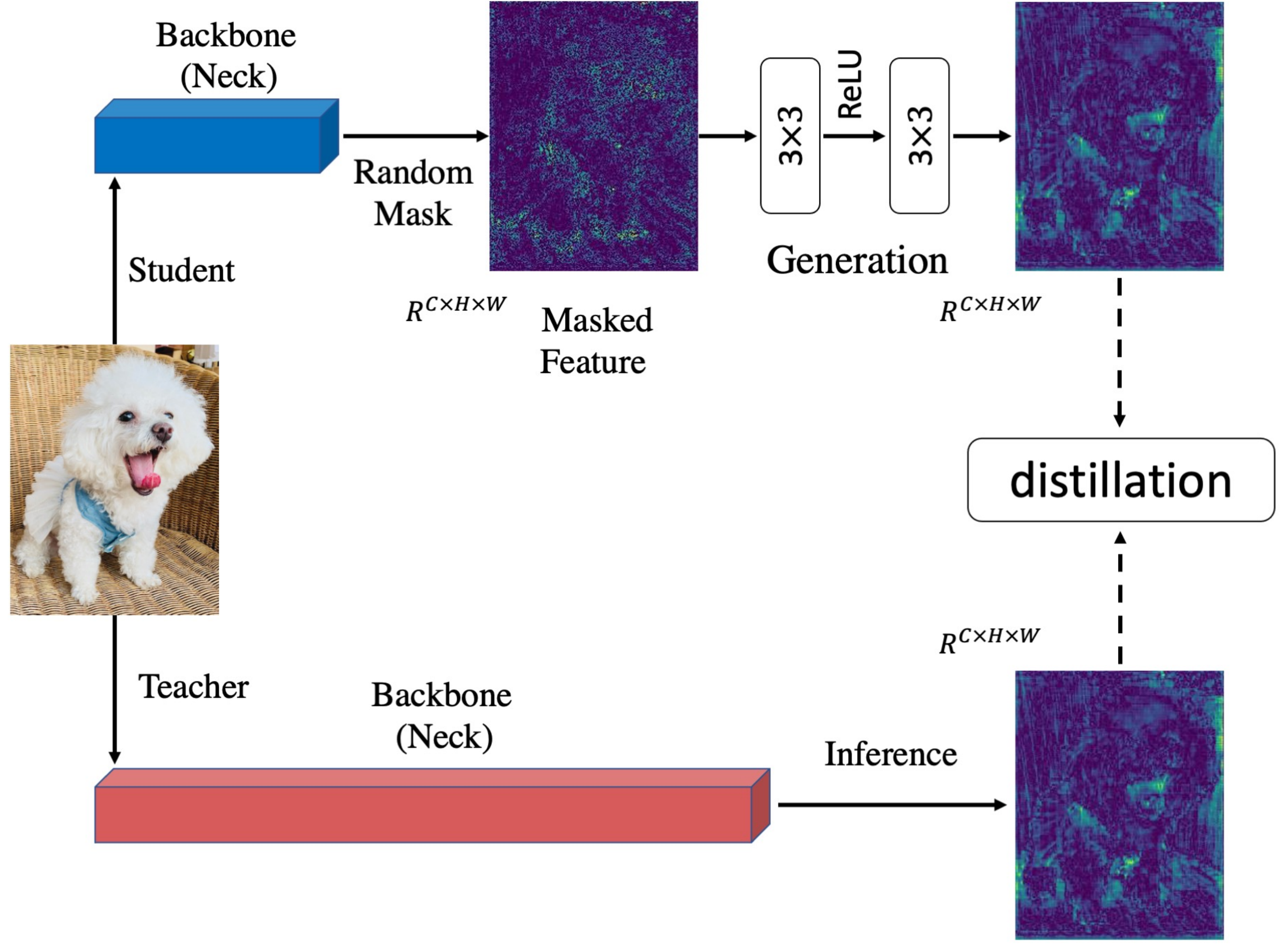}
\caption{An illustration of MGD, Masked Generative Distillation. We first randomly mask the student's feature. And then we use the projector layer to force the student to generate the teacher's feature with the masked feature.}
\label{fig:architecture}
\end{figure}

This method helps the student to mimic the teacher's features directly. However, we propose masked generative distillation (MGD), which aims at forcing the student to generate the teacher's feature instead of mimicking it, bringing the student significant improvements in both classification and dense prediction. The architecture of MGD is shown in  Fig. \ref{fig:architecture} and we will introduce it specifically in this section.

\subsection{Generation with Masked Feature}

For CNN-based models, features of deeper layers have a larger receptive field and better representation of the original input image. In other words, the feature map pixels already contain the information of adjacent pixels to a certain extent. Therefore, we can use partial pixels to recover the complete feature map. Our method aims at generating the teacher's feature by student's masked feature, which can help the student achieve a better representation.

We denote by the $T^{l} \in R^{C\times H\times W}$ and $S^{l}\in R^{C\times H\times W}(l=1,..,L)$ the $l$-th feature map of the teacher and student, respectively. Firstly we set $l$-th random mask to cover the student's $l$-th feature, which can be formulated as:
\begin{equation}
    \label{mask}
    M_{i,j}^{l}=
    \begin{cases}
        0, & \text{if}\ \ R_{i,j}^{l}< \lambda \\ 
        1, & \text{Otherwise}
    \end{cases}
\end{equation}
where $R_{i,j}^{l}$ is a random number in $(0,1)$ and $i, j$ are the the horizontal and vertical coordinates of the feature map, respectively. $\lambda$ is a hyper-parameter that denotes the masked ratio. The $l-th$ feature map is covered by the $l$-th random mask.

Then we use the corresponding mask to cover the student's feature map and try to generate teacher's feature maps with the left pixels, which can be formulated as follows:
\begin{equation}
    \mathcal{G}\big(f_{align}(S^{l})\cdot M^{l}\big)\longrightarrow T^{l}
  \label{generation method}
\end{equation}
\begin{equation}
    \mathcal{G}(F) = W_{l2}(ReLU(W_{l1}(F)))
  \label{projector}
\end{equation}
$\mathcal{G}$ denotes the projector layer which includes two convolutional layers: $W_{l1}$ and $W_{l2}$, one activation layer $ReLU$. In this paper, we adopt 1$\times$1 convolutional layers for the adaptation layer $f_{align}$, 3$\times$3 convolutional layers for projector layer $W_{l1}$ and $W_{l2}$.

According to this method, we design the distillation loss $L_{dis}$ for MGD:
\begin{equation}
    L_{dis}(S,T)=\sum_{l=1}^{L}\sum_{k=1}^{C}\sum_{i=1}^{H}\sum_{j=1}^{W}\Big( T_{k,i,j}^{l}-\mathcal{G}\big(f_{align}(S_{k,i,j}^{l})\cdot M_{i,j}^{l}\big)\Big)^{2}
  \label{mgd_loss}
\end{equation}
where $L$ is the sum of layers for distillation and  $C, H, W$ denote the shape of the feature map. $S$ and $T$ denote the feature of the student and teacher, respectively.

\subsection{Overall Loss}
With the proposed distillation loss $L_{dis}$ for MGD, we train all the models with the total loss as follows:
\begin{equation}
    \label{all loss}
        L_{all}=L_{original}+\alpha\cdot L_{dis}
\end{equation}
where $L_{original}$ is the original loss for the models among all the tasks and $\alpha$ is a hyper-parameter to balance the loss.

MGD is a simple and effective method for distillation and can be applied to various tasks easily. The process of our method is summarized in Algorithm \ref{algo}.

\begin{algorithm}
\caption{Masked Generative Distillation}
\hspace*{0.02in} {\bf Input:}
Teacher: $T$, Student: $S$, Input: $x$, label: $y$, hyper-parameter: $\alpha,\lambda$
\begin{algorithmic}[1]
\State Using $S$ to get the feature $fea^{S}$ and output $\hat{y}$ of Input $x$
\State Using $T$ to get the feature $fea^{T}$ of Input $x$
\State Calculating the original loss of the model: $L_{original}(\hat{y},y)$
\State Calculating the distillation loss in Equation~\ref{mgd_loss}: $L_{dis}(fea^{S},fea^{T})$
\State Using $L_{all} = L_{original}+\alpha\cdot L_{dis}$ to update $S$ 
\end{algorithmic}
\hspace*{0.02in} {\bf Output:}
$S$
\label{algo}
\end{algorithm}

\section{Main Experiments}
MGD is a feature-based distillation that can easily be applied to different models for various tasks. In this paper, we conduct experiments on various tasks, including classification, object detection, semantic segmentation and instance segmentation. We experiment with different models and datasets for different tasks, and all the models achieve excellent improvements with MGD.

\subsection{Classification}
\subsubsection{Datasets.} For classification task, we evaluate our knowledge distillation method on ImageNet\cite{deng2009imagenet}, which contains 1000 object categories. We use the 1.2 million images for training and 50k images for testing for all the classification experiments. We use accuracy to evaluate the models.

\subsubsection{Implementation Details.} For the classification task, we calculate the distillation loss on the last feature map from the backbone. The ablation study about this is shown in Section~\ref{sec:stages}. MGD uses a hyper-parameter $\alpha$ to balance the distillation loss in Equation~\ref{all loss}. The other hyper-parameter $\lambda$ is used to adjust the masked ratio in Equation~\ref{mask}. We adopt the hyper-parameters $\{\alpha = 7 \times 10^{-5}, \lambda=0.5\}$ for all the classification experiments. We train all the models for 100 epochs with SGD optimizer, where the momentum is 0.9 and the weight decay is 0.0001. We initialize the learning rate to 0.1 and decay it for every 30 epochs. This setting is based on 8 GPUs. The experiments are conducted with MMClassification\cite{2020mmclassification} and MMRazor\cite{2021mmrazor} based on Pytorch\cite{paszke2019pytorch}.

\setlength{\tabcolsep}{3.0pt}
\begin{table*}
  \centering
    \caption{Results of different distillation methods on ImageNet dataset. {\bf T} and {\bf S} mean the teacher and student, respectively.}
  \begin{tabular}{c|l|cc|l|cc}
    \toprule
    Type& Method & Top-1  & Top-5 & Method &Top-1&Top-5\\
    \midrule
    &ResNet-34(T) & 73.62 &91.59&ResNet-50(T)&76.55&93.06\\
    &ResNet-18(S) & 69.90 &89.43&MobileNet(S)&69.21&89.02\\
    \midrule
    \multirow{2}{*}{\makecell{Logit}}
    &KD\cite{hinton2015distilling} & 70.68&90.16&KD\cite{hinton2015distilling}&70.68&90.30\\
    &WSLD\cite{zhou2020rethinking} & 71.54&90.25&WSLD\cite{zhou2020rethinking}&72.02&90.70\\
    \midrule
    \multirow{6}{*}{\makecell{Feature}}
    &AT\cite{zagoruyko2016paying} & 70.59&89.73&AT\cite{zagoruyko2016paying}&70.72&90.03\\
    &OFD\cite{heo2019comprehensive} & 71.08&90.07&OFD\cite{heo2019comprehensive}&71.25&90.34\\
    &RKD\cite{park2019relational} & 71.34&90.37&RKD\cite{park2019relational}&71.32&90.62\\
    &CRD\cite{tian2019contrastive} & 71.17&90.13&CRD\cite{tian2019contrastive}&71.40&90.42\\
    &KR\cite{chen2021distilling} & 71.61&90.51&KR\cite{chen2021distilling}&72.56&{\bf91.00}\\
    &\cellcolor{lightgray!45}Ours & \cellcolor{lightgray!45}71.58&\cellcolor{lightgray!45}90.35&\cellcolor{lightgray!45}Ours&\cellcolor{lightgray!45}72.35&\cellcolor{lightgray!45}90.71\\
    \midrule
    \multirow{2}{*}{\makecell{Feature $+$ Logit}}
    &SRRL\cite{yang2020knowledge} & 71.73&{\bf90.60}&SRRL\cite{yang2020knowledge}&72.49&90.92\\
    &\cellcolor{lightgray!45}Ours+WSLD &\cellcolor{lightgray!45}{\bf71.80}&\cellcolor{lightgray!45}90.40&\cellcolor{lightgray!45}Ours+WSLD&\cellcolor{lightgray!45}{\bf72.59}&\cellcolor{lightgray!45}90.94\\
    \bottomrule
  \end{tabular}
  \label{table:classification results}
\end{table*}

\subsubsection{Classification Results.} We conduct experiments with two popular distillation settings for classification, including homogeneous and heterogeneous distillation. The first distillation setting is from ResNet-34\cite{he2016deep} to ResNet-18, the other setting is from ResNet-50 to MobileNet\cite{howard2017mobilenets}. As shown in Table~\ref{table:classification results}, we compare with various knowledge distillation methods\cite{hinton2015distilling,zagoruyko2016paying,heo2019comprehensive,park2019relational,tian2019contrastive,chen2021distilling,zhou2020rethinking,yang2020knowledge}, including feature-based methods, logit-based methods and the combination. The student ResNet-18 and MobileNet gain 1.68 and 3.14 Top-1 accuracy improvement with our method, respectively. Besides, as described above, MGD just needs to calculate the distillation loss on the feature maps and can be combined with other logit-based methods for image classification. So we try to add the logit-based distillation loss in WSLD\cite{zhou2020rethinking}. In this way, the two students achieve 71.80 and 72.59 Top-1 accuracy, getting another 0.22 and 0.24 improvement, respectively.

\subsection{Object Detection and Instance Segmentation}
\subsubsection{Datasets.} We conduct experiments on COCO2017 dataset\cite{lin2014microsoft}, which contains 80 object categories. We use the 120k train images for training and 5k val images for testing. The performances of models are evaluated in Average Precision.

\begin{table*}
  \centering
    \caption{Results of different distillation methods for object detection on COCO.}
  \begin{tabular}{c|l|lccc}
    \toprule
    Teacher& Student & mAP  & AP$_{S}$ & AP$_{M}$ &AP$_{L}$\\
    \midrule
    \multirow{5}{*}{\makecell{RetinaNet\\ResNeXt101\\(41.0)}}
    &RetinaNet-Res50 & 37.4 &20.6&40.7&49.7\\
    &FKD\cite{zhang2020improve} & 39.6&22.7&43.3&52.5\\
    &CWD\cite{shu2021channel}&40.8&22.7&44.5&55.3\\
    &FGD\cite{yang2021focal}&40.7&22.9&45.0&54.7\\
    &\cellcolor{lightgray!45}Ours & \cellcolor{lightgray!45}{\bf41.0}&\cellcolor{lightgray!45}{\bf23.4}&\cellcolor{lightgray!45}{\bf45.3}&\cellcolor{lightgray!45}{\bf55.7}\\
    \midrule
    \multirow{5}{*}{\makecell{Cascade\\Mask RCNN\\ResNeXt101\\(47.3)}}
    &Faster RCNN-Res50 & 38.4 &21.5&42.1&50.3\\
    &FKD\cite{zhang2020improve} & 41.5&23.5&45.0&55.3\\
    &CWD\cite{shu2021channel}&41.7&23.3&45.5&55.5\\
    &FGD\cite{yang2021focal} &42.0&{\bf23.8}&46.4&55.5\\
    &\cellcolor{lightgray!45}Ours & \cellcolor{lightgray!45}{\bf42.1}&\cellcolor{lightgray!45}23.7&\cellcolor{lightgray!45}{\bf46.4}&\cellcolor{lightgray!45}{\bf56.1}\\
    \midrule
    \multirow{5}{*}{\makecell{RepPoints\\ResNeXt101\\(44.2)}}
    &RepPoints-Res50 & 38.6&22.5&42.2&50.4\\
    &FKD\cite{zhang2020improve} & 40.6&23.4&44.6&53.0\\
    &CWD\cite{shu2021channel}&42.0&24.1&46.1&55.0\\
    &FGD\cite{yang2021focal} &42.0&24.0&45.7&55.6\\
    &\cellcolor{lightgray!45}Ours & \cellcolor{lightgray!45}{\bf42.3}&\cellcolor{lightgray!45}{\bf24.4}&\cellcolor{lightgray!45}{\bf46.2}&\cellcolor{lightgray!45}{\bf55.9}\\
    \bottomrule
  \end{tabular}
  \label{table:detection results}
\end{table*}

\subsubsection{Implementation Details.} We calculate the distillation loss on all the feature maps from the neck. We adopt the hyper-parameters $\{\alpha = 2 \times 10^{-5}, \lambda=0.65\}$ for all the one-stage models, $\{\alpha = 5 \times 10^{-7}, \lambda=0.45\}$ for all the two-stage models. We train all the models with SGD optimizer, where the momentum is 0.9 and the weight decay is 0.0001. Unless specified, we train the models for 24 epochs. We use inheriting strategy\cite{kang2021instance,yang2021focal} which initializes the student with the teacher’s neck and head parameters to train the student when they have the same head structure. The experiments are conducted with MMDetection\cite{mmdetection}.

\subsubsection{Object Detection and Instance Segmentation Results.} For object detection, we conduct experiments on three different types of detectors, including a two-stage detector (Faster RCNN\cite{ren2015faster}), an anchor-based one-stage detector (RetinaNet\cite{lin2017focal}) and an anchor-free one-stage detector (RepPoints\cite{yang2019reppoints}). We compare MGD with three recent state-of-the-art distillation methods for detectors\cite{zhang2020improve,shu2021channel,yang2021focal}. For instance segmentation, we conduct experiments on two models, SOLO\cite{wang2020solo} and Mask RCNN\cite{he2017mask}. As shown in Table~\ref{table:detection results} and Table~\ref{table:instance segmentation results}, our method surpasses the other state-of-the-art methods for both object detection and instance segmentation. The students gain significant AP improvements with MGD, {\em e.g. }the ResNet-50 based RetinaNet and SOLO gets 3.6 Boundingbox mAP and 3.1 Mask mAP improvement on COCO dataset, respectively.

\begin{table*}
  \centering
    \caption{Results of different distillation methods for instance segmentation on COCO. {\bf MS} means multi-scale training. Here the AP means Mask AP.}
  \begin{tabular}{c|l|lccc}
    \toprule
    Teacher &Students &mAP &AP$_{S}$ & AP$_{M}$ & AP$_{L}$\\
    \midrule
    \multirow{3}{*}{\makecell{SOLO-Res101\\3x,MS(37.1)}}
    &SOLO-Res50(1x) & 33.1&12.2&36.1&50.8\\
    &FGD\cite{yang2021focal} & 36.0&14.5&39.5&54.5\\
    &\cellcolor{lightgray!45}Ours & \cellcolor{lightgray!45}{\bf36.2}&\cellcolor{lightgray!45}{\bf14.2}&\cellcolor{lightgray!45}{\bf39.7}&\cellcolor{lightgray!45}{\bf55.3}\\
    \midrule
    \multirow{3}{*}{\makecell{Cascade\\Mask RCNN\\ResNeXt101(41.1)}}
    &Mask RCNN-Res50 & 35.4&16.6&38.2&52.5\\
    &FGD\cite{yang2021focal} & 37.8&17.1&40.7&56.0\\
    &\cellcolor{lightgray!45}Ours & \cellcolor{lightgray!45}{\bf 38.1}&\cellcolor{lightgray!45}{\bf17.1}&\cellcolor{lightgray!45}{\bf41.1}&\cellcolor{lightgray!45}{\bf56.3}\\
    \bottomrule
  \end{tabular}
  \label{table:instance segmentation results}
\end{table*}

\subsection{Semantic Segmentation}
\subsubsection{Datasets.} For the semantic segmentation task, we evaluate our method on CityScapes dataset\cite{cordts2016cityscapes}, which contains 5000 high-quality images (2975, 500, and 1525 images for the training, validation, and testing). We evaluate all the models with mean Intersection-over-Union (mIoU).

\subsubsection{Implementation Details.} For all the models, we calculate the distillation loss on the last feature map from the backbone. We adopt the hyper-parameters $\{\alpha = 2 \times 10^{-5}, \lambda=0.75\}$ for all the experiments. We train all the models with SGD optimizer, where the momentum is 0.9 and the weight decay is 0.0005. We run all the models on 8 GPUs. The experiments are conducted with MMSegmentation\cite{mmseg2020}.

\subsubsection{Semantic Segmentation Results.} For the semantic segmentation task, we conduct experiments on two settings. In both settings, we use PspNet-Res101\cite{zhao2017pyramid} as the teacher and train it for 80k iterations with 512$\times$1024 input size. We use PspNet-Res18 and DeepLabV3-Res18\cite{chen2017rethinking} as students and train them for 40k iterations with 512$\times$1024 input size. As shown in Table~\ref{table:segmentation results}, our method surpasses the state-of-the-art distillation method for semantic segmentation. Both the homogeneous and heterogeneous distillation bring the students significant improvements, {\em e.g. }the ResNet-18 based PspNet gets 3.78 mIoU improvement. Besides, MGD is a feature-based distillation method and can be combined with other logit-based distillation methods. As the results show, the student PspNet and DeepLabV3 get another 0.47 and 0.29 mIoU improvement by adding the logit distillation loss of the head in CWD\cite{shu2021channel}.

\begin{table*}
  \centering
    \caption{Results of the semantic segmentation task on CityScapes dataset. {\bf T} and {\bf S} mean teacher and student, respectively. The results are the average value of three runs. {\bf $^{*}$} means adding the distillation loss of the head in CWD\cite{shu2021channel}}
  \begin{tabular}{l|lc}
    \toprule
     Method & Input Size  & mIoU\\
    \midrule
    PspNet-Res101(T)&$512\times 1024$&78.34\\
    PspNet-Res18(S) & $512\times 512$ &69.85\\
    SKDS\cite{liu2019structured}&$512\times 512$&72.70\\
    CWD\cite{shu2021channel}&$512\times 512$&73.53\\
    \cellcolor{lightgray!45}Ours & \cellcolor{lightgray!45}$512\times 512$&\cellcolor{lightgray!45}73.63\\
    \cellcolor{lightgray!45}Ours$^{*}$ & \cellcolor{lightgray!45}$512\times 512$&\cellcolor{lightgray!45}74.10\\
    \midrule
    PspNet-Res101(T)&$512\times 1024$&78.34\\
    DeepLabV3-Res18(S) & $512\times 512$ &73.20\\
    SKDS\cite{liu2019structured}&$512\times 512$&73.87\\
    CWD\cite{shu2021channel}&$512\times 512$&75.93\\
    \cellcolor{lightgray!45}Ours&\cellcolor{lightgray!45}$512\times 512$&\cellcolor{lightgray!45}76.02\\
    \cellcolor{lightgray!45}Ours$^{*}$ & \cellcolor{lightgray!45}$512\times 512$&\cellcolor{lightgray!45}76.31\\
    \bottomrule
  \end{tabular}

  \label{table:segmentation results}
\end{table*}

\section{Analysis}
\subsection{Better representation with MGD}

MGD forces the student to generate the teacher's complete feature map with its masked feature instead of mimicking it directly. It helps the students get a better representation of the input image. In this subsection, we study this by using the student to teach itself. We first train ResNet-18 directly as a teacher and the baseline. Then we use the trained ResNet-18 to distill itself with MGD. For comparison, we also distill the student by forcing the student to mimic the teacher directly. The distillation loss for mimicking is the square of L2 distance between the student's feature map and the teacher's feature map.

As shown in Table~\ref{table:self dis}, the student also gains 1.01 accuracy improvement with MGD even when the teacher is itself. In contrast, the improvement is very limited when forcing the student to mimic the teacher's feature map directly. The comparison indicates that the student's feature map achieves better representation than the teacher's after distillation.

\begin{figure}[!ht]
\centering
\subfigure[Distillation by MGD]{
\includegraphics[{width=5.35cm}]{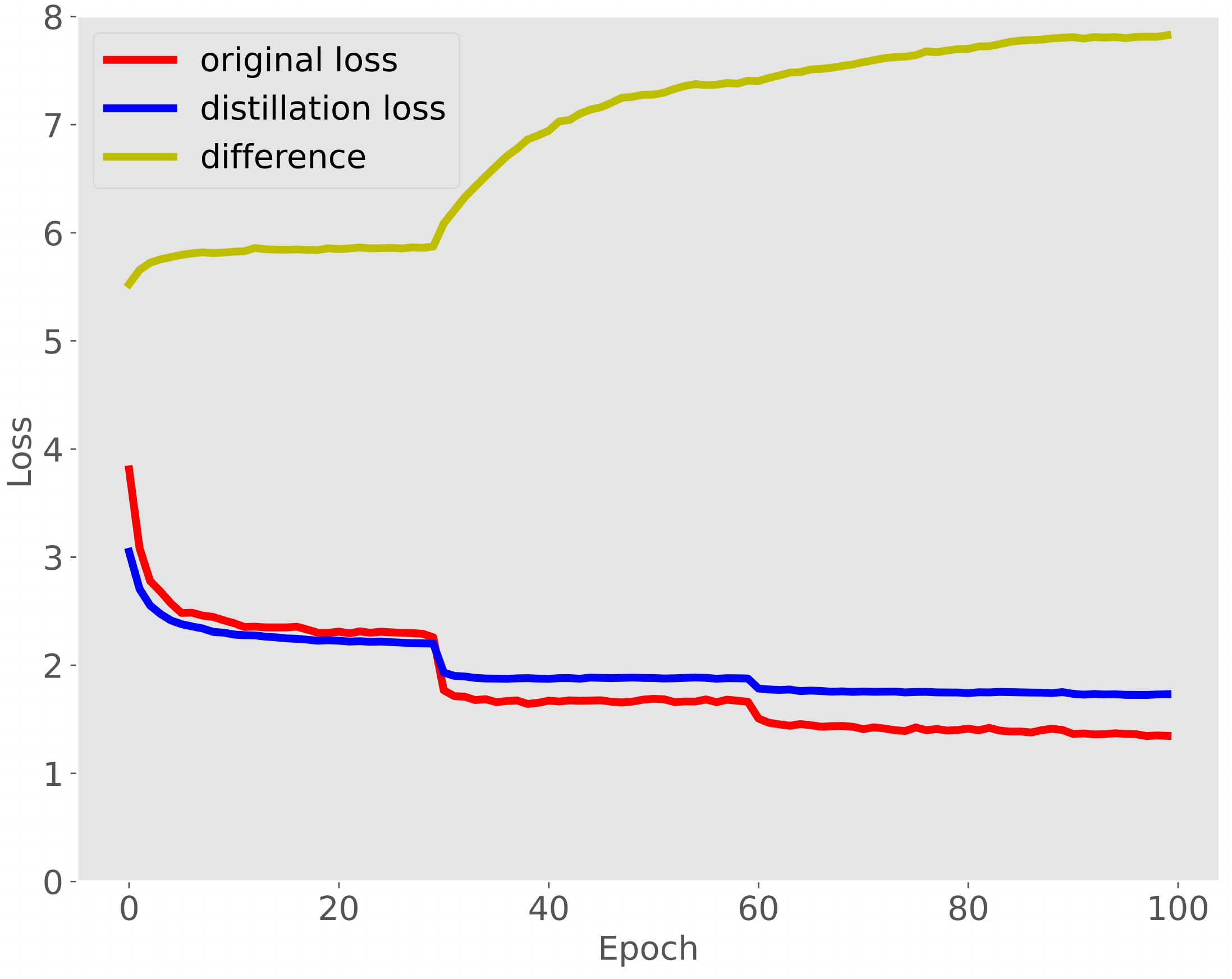}
\label{fig:mgd loss}}
\quad
\subfigure[Distillation by mimicking the teacher]{
\includegraphics[width=5.35cm]{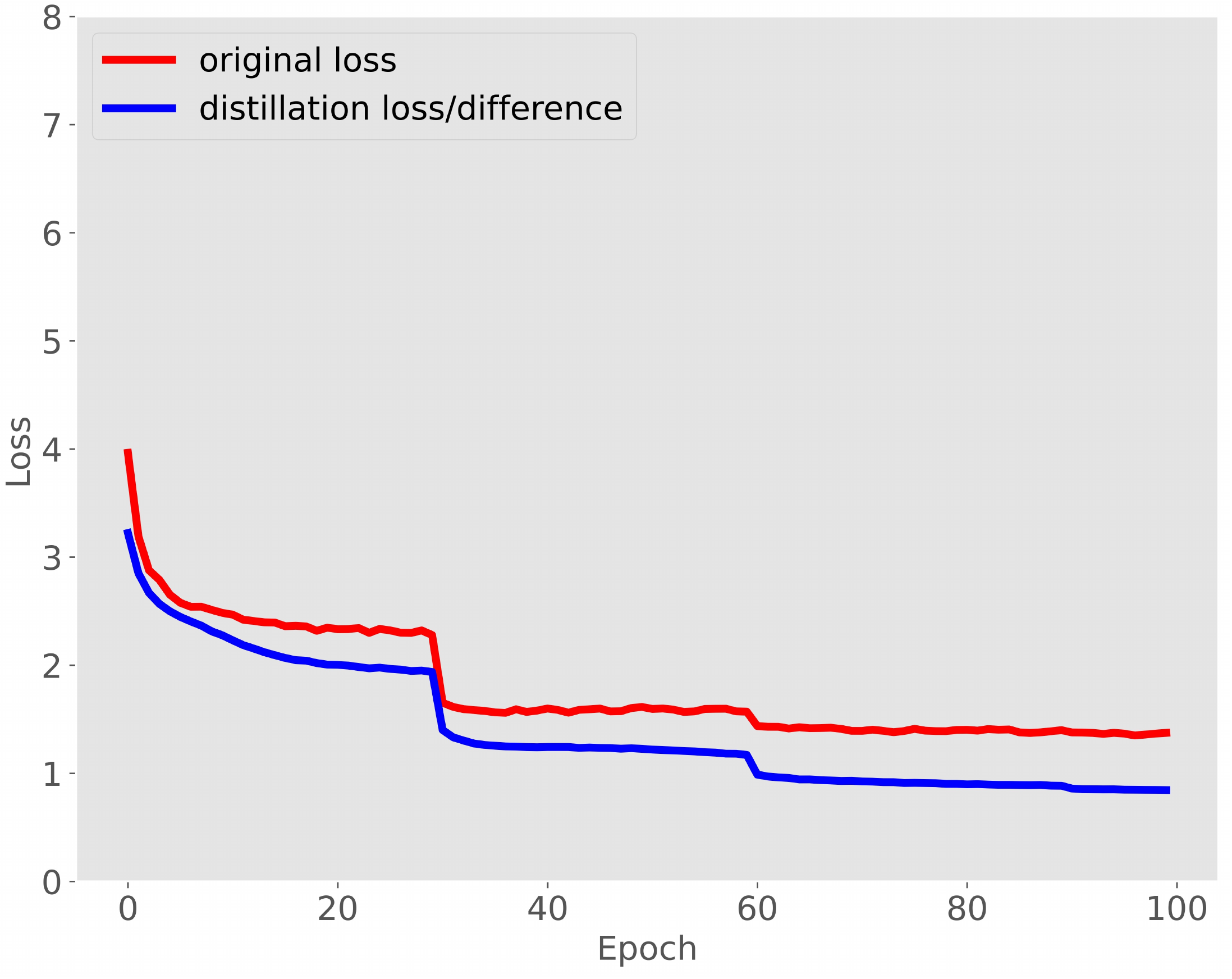}
\label{fig:mimic loss}}
\caption{The training loss curves of ResNet-18 distilling ResNet-18. {\bf Difference} means the square of L2 distance between the last feature map of student and teacher. It is also the distillation loss for mimicking the teacher.}
\label{fig:self dis}
\end{figure}

Furthermore, we visualize the training loss curves for distillation with MGD and mimicking the teacher, which is shown in Figure~\ref{fig:self dis}. The {\bf difference} in the figure means the square of L2 distance between the last feature map of student and teacher, which is also the distillation loss for mimicking the teacher. As the figure shows, the {\bf difference} keeps decreasing during mimicking the teacher directly and finally the student gets a similar feature to the teacher. However, the improvement with this method is minimal. In contrast, the {\bf difference} becomes larger after training with MGD. Although the student gets a different feature from the teacher, it gets higher accuracy, also indicating the student's feature obtains stronger representation power.

\begin{table*}
  \centering
    \caption{The results of distillation for Rse18-18 on ImageNet. We train ResNet-18 directly as the teacher and student baseline. {\bf T} and {\bf S} mean teacher and student.}
  \begin{tabular}{l|lc}
    \toprule
      & Top-1  & Top-5\\
    \midrule
    ResNet-18(T,S)&69.90&89.43\\
    +mimicking feature&70.05&89.41\\
    +MGD&70.91&89.82\\
    \bottomrule
  \end{tabular}
  \label{table:self dis}
\end{table*}

\subsection{Distillation by masking random channels}
For image classification, the models usually utilize a pooling layer to reduce the spatial dimension of the feature map. This layer makes the model more sensitive to the channels than spatial pixels. So in this subsection, we try to apply MGD by masking random channels instead of spatial pixels for classification. We adopt the masked ratio $\beta=0.15$ and hyper-parameter $\alpha=7\times10^{-5}$ for the experiments.  As shown in Table~\ref{table:channel results}, the student can get better performance by masking random channels instead of spatial for image classification. The student Res-18 and MobileNet achieve 0.13 and 0.14 Top-1 accuracy gains, respectively.
\setlength{\tabcolsep}{3.0pt}
\begin{table}
\begin{center}
\caption{Results of masking random channels on ImageNet dataset.}
\label{table:channel results}
\begin{tabular}{c|c|cc}
\hline
 &Acc & MGD(Spatial)&MGD(Channel)\\
\hline
\multirow{2}{*}{Res34-18}
&Top-1&71.58&{\bf71.69}\\
&Top-5&90.35&{\bf90.42}\\
\hline
\multirow{2}{*}{Res50-mv1}
&Top-1&72.35&{\bf72.49}\\
&Top-5&90.71&{\bf90.94}\\
\hline
\end{tabular}
\end{center}
\end{table}

\subsection{Distillation with different teachers}

\begin{figure}[!ht]
\centering
\includegraphics[{width=0.9\linewidth}]{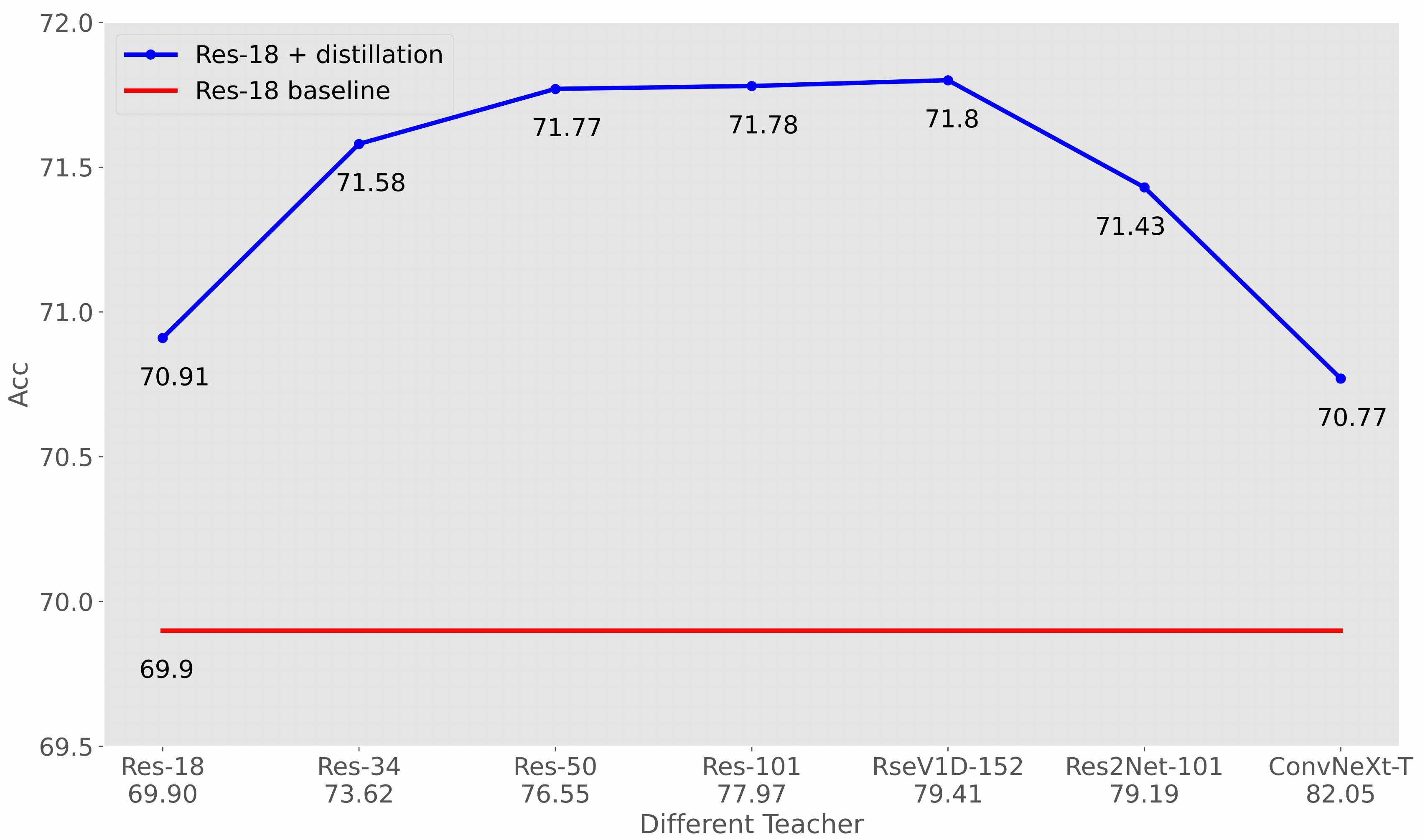}
\caption{The results of training ResNet-18 with our method by different teachers on ImageNet dataset.}
\label{fig:teacher dif}
\end{figure}

Cho et al.\cite{cho2019efficacy} show the teacher with higher accuracy may not be the better teacher for knowledge distillation on image classification. The conclusion is based on the logit-based distillation method. However, our method just needs to calculate the distillation loss on the feature maps. In this subsection, we study this conclusion by using different kinds of teachers to distill the same student ResNet-18, which is shown in Figure~\ref{fig:teacher dif}. 

As shown in Figure~\ref{fig:teacher dif}, the better teacher benefits more to the student when the teacher and student have a similar architecture, {\em e.g.} ResNet-18 achieves 70.91 and 71.8 accuracy with ResNet-18 and ResNetV1D-152 as the teacher, respectively. However, when the teacher and student have a different architecture, it is hard for the student to generate the teacher's feature map and the improvement by distillation is limited. Moreover, the distillation performs worse with a larger difference between the architectures. For example, although Res2Net-101\cite{gao2019res2net} and ConvNeXt-T\cite{liu2022convnet} has 79.19 and 82.05 accuracy, they just bring 1.53 and 0.88 accuracy improvement to the student, which is even lower than the ResNet-34 based teacher(73.62 accuracy).

The results in Figure~\ref{fig:teacher dif} indicate that the stronger teacher is better for feature-based distillation when they have similar architecture. Besides, the homogeneous teacher is much better for feature-based distillation than the teacher with high accuracy but a heterogeneous architecture.

\subsection{The generative block}

MGD uses a simple block to restore the feature, called the generative block. In Equation~\ref{projector}, we use two 3$\times$3 convolutional layers and one activation layer $ReLU$ to accomplish this. In this subsection, we explore the effects of the generative block with different compositions, shown in Table~\ref{table:projector}.

As the results show, the student gets the slightest improvement when there is only one convolutional layer. However, when there are three convolutional layers, the student gets a worse Top-1 but better Top-5 accuracy. As for the kernel size, 5$\times$5 convolutional kernel needs more compute resources, while it gets worse performance. Based on the results, we choose the architecture in Equation~\ref{projector} for MGD, which includes two convolutional layers and one activation layer.
\begin{table*}
  \centering
    \caption{The results of distillation about generative part. The {\bf Conv Layers} mean the sum of convolutional layers and the {\bf kernel size} belongs to the convolutional layer. We add one activation layer $ReLU$ between every two convolutional layers.}
  \begin{tabular}{c|c|lc}
    \toprule
      Conv Layers&kernel size& Top-1  & Top-5\\
    \midrule
    1&3$\times$ 3&71.28&90.30\\
    2&3$\times$ 3&{\bf 71.58}&90.35\\
    3&3$\times$ 3&71.49&{\bf 90.44}\\
    \midrule
    2&5$\times$ 5&71.32&90.28\\
    \bottomrule
  \end{tabular}
  \label{table:projector}
\end{table*}

\subsection{Distillation on different stages}
\label{sec:stages}
Our method can also be applied at other stages of the model. In this subsection, we explore distillation at different stages on ImageNet. We calculate the distillation loss on the corresponding layers of teacher and student. As shown in Table~\ref{table:stage}, distilling the shallower layers is also helpful to the student but very limited. While distilling the deeper stage which contains more semantic information benefits the student more. Furthermore, the features from the early stages are not directly used for classification. Therefore, Distilling such features with the last stage feature together may hurt the student's accuracy.

\begin{table*}
  \centering
    \caption{The results of distillation about different stages for Rse34-18 on ImageNet.}
  \begin{tabular}{l|lc}
    \toprule
      & Top-1  & Top-5\\
    \midrule
    stage 1&70.09&89.40\\
    stage 2&70.21&89.38\\
    stage 3&70.37&89.42\\
    stage 4&{\bf 71.58}&{\bf 90.35}\\
    \midrule
    stage 2+3+4&71.47&90.31\\
    \bottomrule
  \end{tabular}
  \label{table:stage}
\end{table*}

\subsection{Sensitivity study of hyper-parameters}
\begin{figure}[!ht]
\centering
\subfigure{
\includegraphics[{width=5.65cm}]{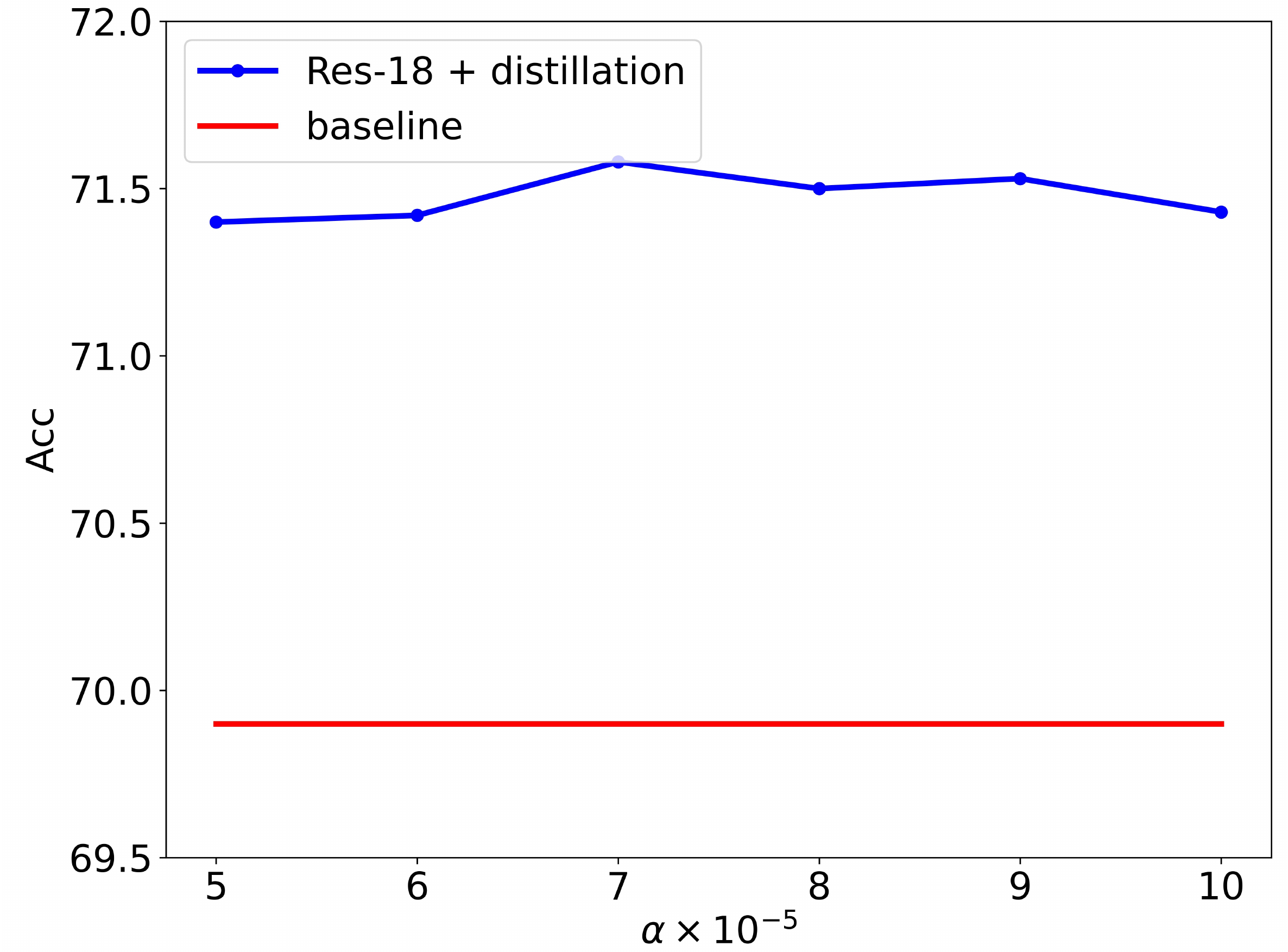}
\label{fig:sens alpha}}
\quad
\subfigure{
\includegraphics[width=5.65cm]{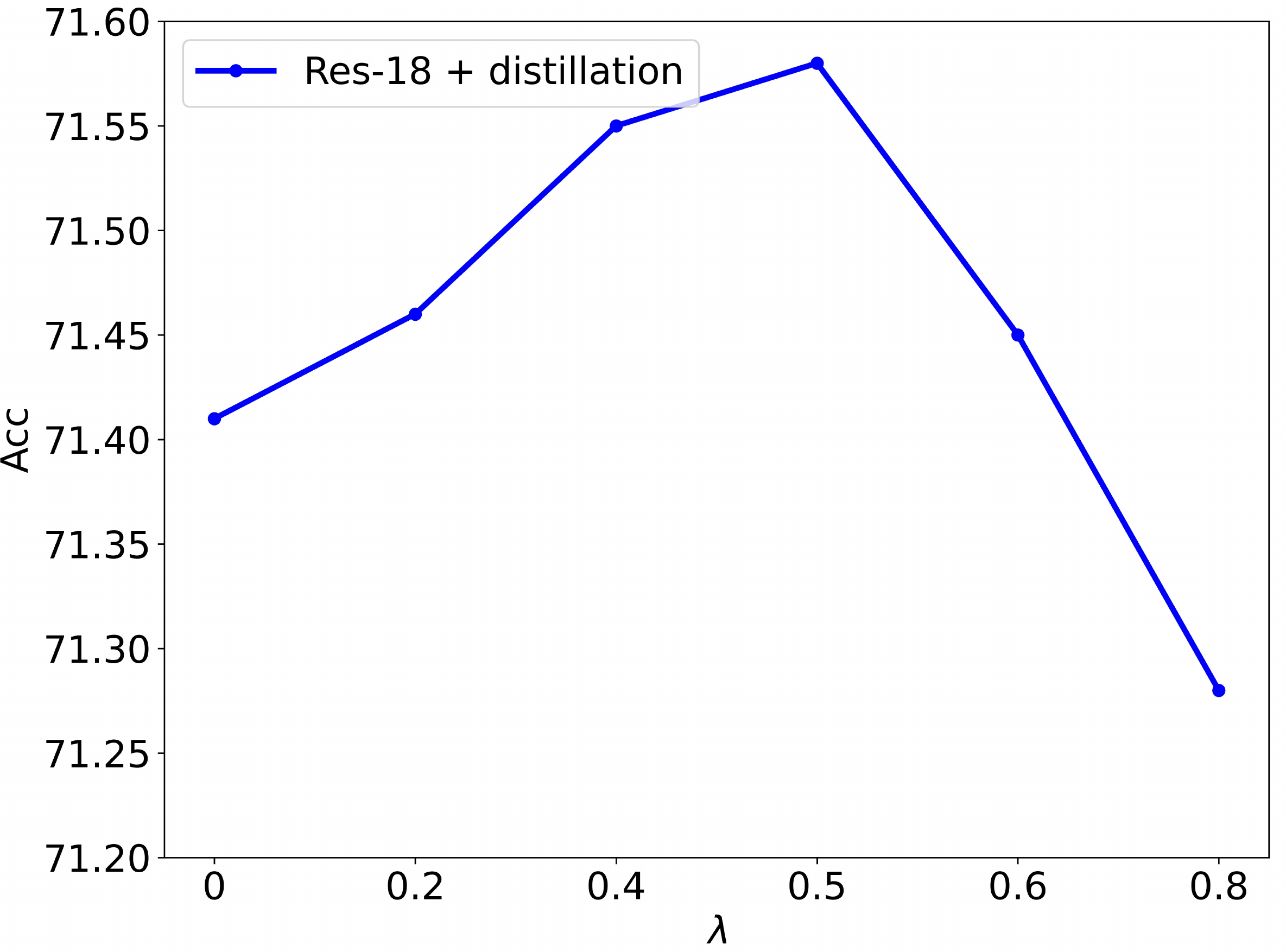}
\label{fig:sens mask}}
\caption{Sensitivity study of hyper-parameters $\alpha$ and $\lambda$ with ResNet34-ResNet18}
\label{fig:sens}
\end{figure}

In this paper, we use $\alpha$ in Equation~\ref{all loss} and $\lambda$ in Equation~\ref{mask} to balance the distillation loss and adjust the mask ratio, respectively. In this subsection, we do the sensitivity study of the hyper-parameters by using ResNet-34 to distill ResNet-18 on ImageNet dataset. The results are shown in Figure~\ref{fig:sens}.

As shown in Figure~\ref{fig:sens}, MGD is not sensitive to the hyper-parameter $\alpha$ which is just used for balancing the loss. As for the mask ratio $\lambda$, the accuracy is 71.41 when it is 0, which means there are no masked parts for the generation. The student gets higher performance with larger ratio when $\lambda < 0.5$. However, when $\lambda$ is too large, {\em e.g.} 0.8, the left semantic information is too poor to generate teacher's complete feature map and the performance improvement is also affected. 

\section{Conclusions}

In this paper, we propose a new way for knowledge distillation, which forces the student to generate the teacher's feature with its masked feature instead of mimicking it directly. Based on this way, we propose a new knowledge distillation method, Masked Generative Distillation (MGD). The students can obtain stronger representation power with MGD. Furthermore, our method is just based on the feature map so that MGD can easily be applied to various tasks such as image classification, object detection, semantic segmentation, and instance segmentation. Extensive experiments on various models with different datasets prove that our method is simple and efficient.

{\bf Acknowledgement.} This work was supported by the SZSTC project Grant No.JCYJ20190809172201639 and Grant No.WDZC20200820200655001, Shenzhen Key Laboratory ZDSYS20210623092001004.

\clearpage
%
%
\bibliographystyle{splncs04}
\bibliography{egbib}

\begin{thebibliography}{10}
\providecommand{\url}[1]{\texttt{#1}}
\providecommand{\urlprefix}{URL }
\providecommand{\doi}[1]{https://doi.org/#1}

\bibitem{chen2017learning}
Chen, G., Choi, W., Yu, X., Han, T., Chandraker, M.: Learning efficient object
  detection models with knowledge distillation. Advances in neural information
  processing systems  \textbf{30} (2017)

\bibitem{mmdetection}
Chen, K., Wang, J., Pang, J., Cao, Y., Xiong, Y., Li, X., Sun, S., Feng, W.,
  Liu, Z., Xu, J., Zhang, Z., Cheng, D., Zhu, C., Cheng, T., Zhao, Q., Li, B.,
  Lu, X., Zhu, R., Wu, Y., Dai, J., Wang, J., Shi, J., Ouyang, W., Loy, C.C.,
  Lin, D.: {MMDetection}: Open mmlab detection toolbox and benchmark. arXiv
  preprint arXiv:1906.07155  (2019)

\bibitem{chen2017rethinking}
Chen, L.C., Papandreou, G., Schroff, F., Adam, H.: Rethinking atrous
  convolution for semantic image segmentation. arXiv preprint arXiv:1706.05587
  (2017)

\bibitem{chen2021distilling}
Chen, P., Liu, S., Zhao, H., Jia, J.: Distilling knowledge via knowledge
  review. In: Proceedings of the IEEE/CVF Conference on Computer Vision and
  Pattern Recognition. pp. 5008--5017 (2021)

\bibitem{cho2019efficacy}
Cho, J.H., Hariharan, B.: On the efficacy of knowledge distillation. In:
  Proceedings of the IEEE/CVF international conference on computer vision. pp.
  4794--4802 (2019)

\bibitem{2020mmclassification}
Contributors, M.: Openmmlab's image classification toolbox and benchmark.
  \url{https://github.com/open-mmlab/mmclassification} (2020)

\bibitem{2021mmrazor}
Contributors, M.: Openmmlab model compression toolbox and benchmark.
  \url{https://github.com/open-mmlab/mmrazor} (2021)

\bibitem{mmseg2020}
Contributors, M.: {MMSegmentation}: Openmmlab semantic segmentation toolbox and
  benchmark. \url{https://github.com/open-mmlab/mmsegmentation} (2020)

\bibitem{cordts2016cityscapes}
Cordts, M., Omran, M., Ramos, S., Rehfeld, T., Enzweiler, M., Benenson, R.,
  Franke, U., Roth, S., Schiele, B.: The cityscapes dataset for semantic urban
  scene understanding. In: Proceedings of the IEEE conference on computer
  vision and pattern recognition. pp. 3213--3223 (2016)

\bibitem{dai2021general}
Dai, X., Jiang, Z., Wu, Z., Bao, Y., Wang, Z., Liu, S., Zhou, E.: General
  instance distillation for object detection. In: Proceedings of the IEEE/CVF
  Conference on Computer Vision and Pattern Recognition. pp. 7842--7851 (2021)

\bibitem{deng2009imagenet}
Deng, J., Dong, W., Socher, R., Li, L.J., Li, K., Fei-Fei, L.: Imagenet: A
  large-scale hierarchical image database. In: 2009 IEEE conference on computer
  vision and pattern recognition. pp. 248--255. Ieee (2009)

\bibitem{gao2019res2net}
Gao, S.H., Cheng, M.M., Zhao, K., Zhang, X.Y., Yang, M.H., Torr, P.: Res2net: A
  new multi-scale backbone architecture. IEEE TPAMI  (2021).
  \doi{10.1109/TPAMI.2019.2938758}

\bibitem{guo2021distilling}
Guo, J., Han, K., Wang, Y., Wu, H., Chen, X., Xu, C., Xu, C.: Distilling object
  detectors via decoupled features. In: Proceedings of the IEEE/CVF Conference
  on Computer Vision and Pattern Recognition. pp. 2154--2164 (2021)

\bibitem{he2017mask}
He, K., Gkioxari, G., Doll{\'a}r, P., Girshick, R.: Mask r-cnn. In: Proceedings
  of the IEEE international conference on computer vision. pp. 2961--2969
  (2017)

\bibitem{he2016deep}
He, K., Zhang, X., Ren, S., Sun, J.: Deep residual learning for image
  recognition. In: Proceedings of the IEEE conference on computer vision and
  pattern recognition. pp. 770--778 (2016)

\bibitem{he2019knowledge}
He, T., Shen, C., Tian, Z., Gong, D., Sun, C., Yan, Y.: Knowledge adaptation
  for efficient semantic segmentation. In: Proceedings of the IEEE/CVF
  Conference on Computer Vision and Pattern Recognition. pp. 578--587 (2019)

\bibitem{heo2019comprehensive}
Heo, B., Kim, J., Yun, S., Park, H., Kwak, N., Choi, J.Y.: A comprehensive
  overhaul of feature distillation. In: Proceedings of the IEEE/CVF
  International Conference on Computer Vision. pp. 1921--1930 (2019)

\bibitem{hinton2015distilling}
Hinton, G., Vinyals, O., Dean, J., et~al.: Distilling the knowledge in a neural
  network. arXiv preprint arXiv:1503.02531  \textbf{2}(7) (2015)

\bibitem{howard2017mobilenets}
Howard, A.G., Zhu, M., Chen, B., Kalenichenko, D., Wang, W., Weyand, T.,
  Andreetto, M., Adam, H.: Mobilenets: Efficient convolutional neural networks
  for mobile vision applications. arXiv preprint arXiv:1704.04861  (2017)

\bibitem{kang2021instance}
Kang, Z., Zhang, P., Zhang, X., Sun, J., Zheng, N.: Instance-conditional
  knowledge distillation for object detection. In: In Proc. of the Thirty-Fifth
  Conference on Neural Information Processing Systems (NeurIPS) (2021)

\bibitem{lin2017focal}
Lin, T.Y., Goyal, P., Girshick, R., He, K., Doll{\'a}r, P.: Focal loss for
  dense object detection. In: Proceedings of the IEEE international conference
  on computer vision. pp. 2980--2988 (2017)

\bibitem{lin2014microsoft}
Lin, T.Y., Maire, M., Belongie, S., Hays, J., Perona, P., Ramanan, D.,
  Doll{\'a}r, P., Zitnick, C.L.: Microsoft coco: Common objects in context. In:
  European conference on computer vision. pp. 740--755. Springer (2014)

\bibitem{liu2019structured}
Liu, Y., Chen, K., Liu, C., Qin, Z., Luo, Z., Wang, J.: Structured knowledge
  distillation for semantic segmentation. In: Proceedings of the IEEE/CVF
  Conference on Computer Vision and Pattern Recognition. pp. 2604--2613 (2019)

\bibitem{liu2022convnet}
Liu, Z., Mao, H., Wu, C.Y., Feichtenhofer, C., Darrell, T., Xie, S.: A convnet
  for the 2020s. arXiv preprint arXiv:2201.03545  (2022)

\bibitem{park2019relational}
Park, W., Kim, D., Lu, Y., Cho, M.: Relational knowledge distillation. In:
  Proceedings of the IEEE/CVF Conference on Computer Vision and Pattern
  Recognition. pp. 3967--3976 (2019)

\bibitem{paszke2019pytorch}
Paszke, A., Gross, S., Massa, F., Lerer, A., Bradbury, J., Chanan, G., Killeen,
  T., Lin, Z., Gimelshein, N., Antiga, L., et~al.: Pytorch: An imperative
  style, high-performance deep learning library. Advances in neural information
  processing systems  \textbf{32} (2019)

\bibitem{ren2015faster}
Ren, S., He, K., Girshick, R., Sun, J.: Faster r-cnn: Towards real-time object
  detection with region proposal networks. Advances in neural information
  processing systems  \textbf{28} (2015)

\bibitem{romero2014fitnets}
Romero, A., Ballas, N., Kahou, S.E., Chassang, A., Gatta, C., Bengio, Y.:
  Fitnets: Hints for thin deep nets. arXiv preprint arXiv:1412.6550  (2014)

\bibitem{shu2021channel}
Shu, C., Liu, Y., Gao, J., Yan, Z., Shen, C.: Channel-wise knowledge
  distillation for dense prediction. In: Proceedings of the IEEE/CVF
  International Conference on Computer Vision. pp. 5311--5320 (2021)

\bibitem{tian2019contrastive}
Tian, Y., Krishnan, D., Isola, P.: Contrastive representation distillation. In:
  International Conference on Learning Representations (2019)

\bibitem{wang2019distilling}
Wang, T., Yuan, L., Zhang, X., Feng, J.: Distilling object detectors with
  fine-grained feature imitation. In: Proceedings of the IEEE/CVF Conference on
  Computer Vision and Pattern Recognition. pp. 4933--4942 (2019)

\bibitem{wang2020solo}
Wang, X., Kong, T., Shen, C., Jiang, Y., Li, L.: Solo: Segmenting objects by
  locations. In: European Conference on Computer Vision. pp. 649--665. Springer
  (2020)

\bibitem{yang2020knowledge}
Yang, J., Martinez, B., Bulat, A., Tzimiropoulos, G.: Knowledge distillation
  via softmax regression representation learning. In: International Conference
  on Learning Representations (2020)

\bibitem{yang2019reppoints}
Yang, Z., Liu, S., Hu, H., Wang, L., Lin, S.: Reppoints: Point set
  representation for object detection. In: Proceedings of the IEEE/CVF
  International Conference on Computer Vision. pp. 9657--9666 (2019)

\bibitem{yang2021focal}
Yang, Z., Li, Z., Jiang, X., Gong, Y., Yuan, Z., Zhao, D., Yuan, C.: Focal and
  global knowledge distillation for detectors. arXiv preprint arXiv:2111.11837
  (2021)

\bibitem{zagoruyko2016paying}
Zagoruyko, S., Komodakis, N.: Paying more attention to attention: Improving the
  performance of convolutional neural networks via attention transfer. arXiv
  preprint arXiv:1612.03928  (2016)

\bibitem{zhang2020improve}
Zhang, L., Ma, K.: Improve object detection with feature-based knowledge
  distillation: Towards accurate and efficient detectors. In: International
  Conference on Learning Representations (2020)

\bibitem{zhao2017pyramid}
Zhao, H., Shi, J., Qi, X., Wang, X., Jia, J.: Pyramid scene parsing network.
  In: Proceedings of the IEEE conference on computer vision and pattern
  recognition. pp. 2881--2890 (2017)

\bibitem{zhixing2021distilling}
Zhixing, D., Zhang, R., Chang, M., Liu, S., Chen, T., Chen, Y., et~al.:
  Distilling object detectors with feature richness. Advances in Neural
  Information Processing Systems  \textbf{34} (2021)

\bibitem{zhou2020rethinking}
Zhou, H., Song, L., Chen, J., Zhou, Y., Wang, G., Yuan, J., Zhang, Q.:
  Rethinking soft labels for knowledge distillation: A bias--variance tradeoff
  perspective. In: International Conference on Learning Representations (2020)

\end{thebibliography}
\end{document}